\documentclass[10pt,twocolumn,letterpaper]{article}

\usepackage{cvpr}
\usepackage{times}
\usepackage{epsfig}
\usepackage{graphicx}
\usepackage{amsmath}
\usepackage{amssymb}
\usepackage{enumitem}
\usepackage{indentfirst}
\usepackage{booktabs}
\usepackage[ruled]{algorithm2e}
\usepackage{algpseudocode}
\usepackage{algorithmicx}
\usepackage{amsmath}
\usepackage{amssymb}
\usepackage{multirow}

\DeclareMathOperator*{\argmin}{argmin}

\usepackage[pagebackref=true,breaklinks=true,letterpaper=true,colorlinks,bookmarks=false]{hyperref}

\cvprfinalcopy 

\def\cvprPaperID{7644} 

\begin{document}

\title{ADWPNAS: Architecture-Driven Weight Prediction for Neural Architecture Search}

\author{
Xu Zhang\\
SYSU\\
{\tt\small zhangx629@mail2.sysu.edu.cn}
\and
Junzhou Chen\\
SYSU\\
{\tt\small chenjunzhou@mail.sysu.edu.cn}
\and
Bo Gu\dag\\
SYSU\\
{\tt\small gubo@mail.sysu.edu.cn}
}
\maketitle
\footnote{\dag Corresponding author.}

\begin{abstract}

How to discover and evaluate the true strength of models quickly and accurately is one of the key challenges in Neural Architecture Search (NAS). To cope with this problem, we propose an \textbf{A}rchitecture-\textbf{D}riven \textbf{W}eight \textbf{P}rediction (ADWP) approach for neural architecture search (NAS). 
In our approach, we first design an architecture-intensive search space and then train a HyperNetwork by inputting stochastic encoding architecture parameters. In the trained HyperNetwork, weights of convolution kernels can be well predicted for neural architectures in the search space. Consequently, the target architectures can be evaluated efficiently without any finetuning, thus enabling us to search for the optimal architecture in the space of general networks (macro-search). Through real experiments, we evaluate the performance of the models discovered by the proposed ADWPNAS and results show that one search procedure can be completed in 4.0 GPU hours on CIFAR-10. Moreover, the discovered model obtains a test error of 2.41\% with only 1.52M parameters which is superior to the best existing models.

\begin{figure}[t]

\begin{center}
\includegraphics[width=1\linewidth]{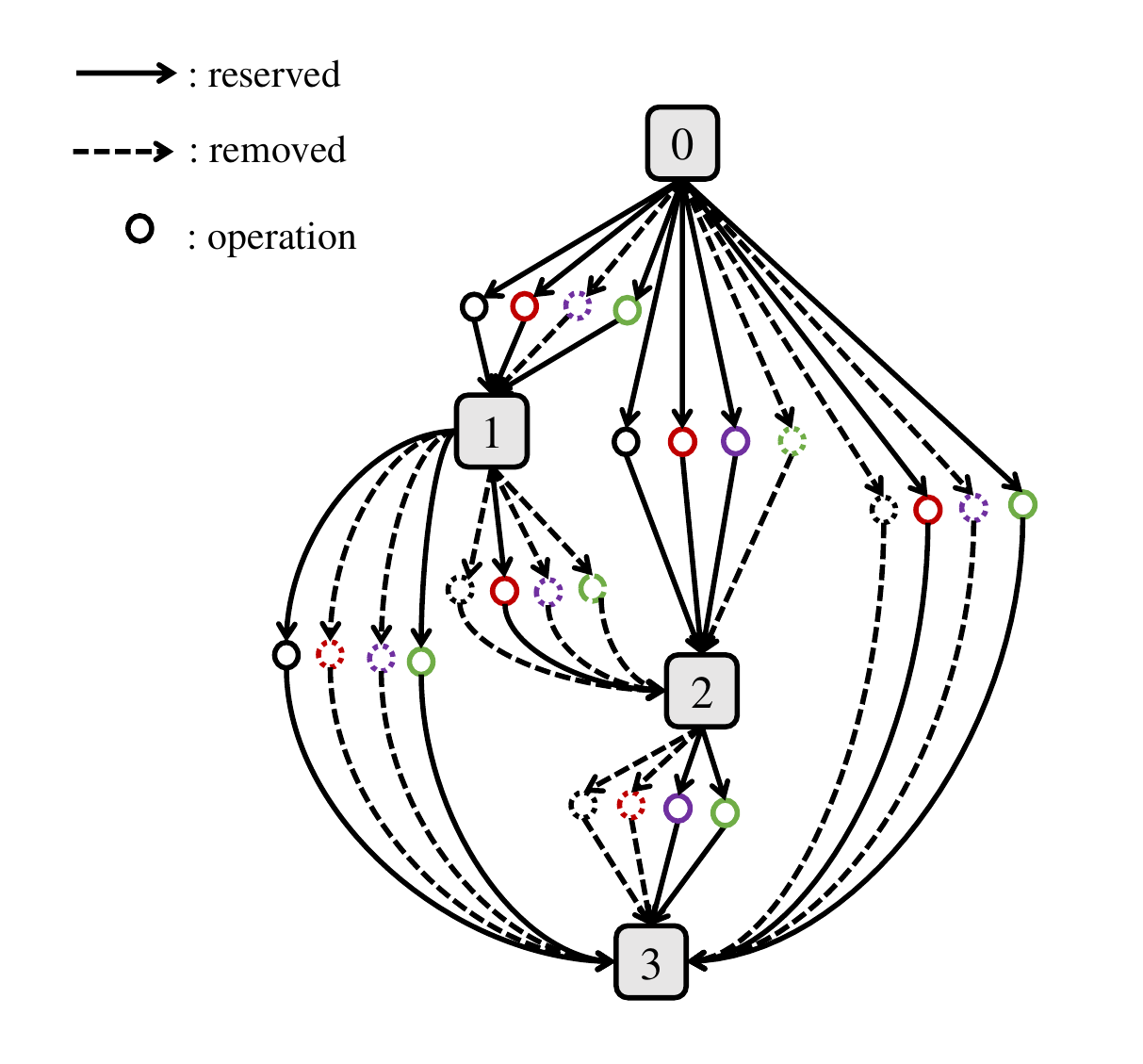}
\end{center}
    \caption{
        The overall of the intensive-space. 
        We represent the search space of each cell as a DAG with ordered nodes. Different operations (colored circles) transform one node (gray square) to intermediate features in a predetermined direction (black arrow). Meanwhile, each node is the sum of the intermediate features transformed from the previous nodes. 
        We prune the search space into an intensive-space as described in Sec.~\ref{subsect:intensive-space}.
        The solid arrows indicate the reserved operations after pruning, and the dotted ones mean the removed operations.
    }
\label{fig:space}
\end{figure}

\end{abstract}

\section{Introduction}

Designing efficient and effective neural architectures has always been of vital importance for deep learning. In recent years, Neural Architecture Search (NAS)~\cite{zoph2016neural,real2017large,zoph2018learning,pham2018efficient,real2019regularized,liu2018darts} has demonstrated superior capabilities in discovering excellent neural architectures automatically. Specifically, neural architectures obtained through NAS methods achieve outstanding performance on the tasks of computer vision, such as image classification~\cite{chen2019progressive}, object detection~\cite{ghiasi2019fpn} and semantic segmentation~\cite{liu2019auto}.

Most NAS methods rely on reinforcement learning (RL)~\cite{zoph2016neural,zoph2018learning,bello2017neural} or evolutionary algorithms (EA)~\cite{real2019regularized,real2017large,liu2017hierarchical}, which incurs intensive computation during the search procedures~\cite{zoph2018learning,real2019regularized}. For instance, a RL-based method ~\cite{zoph2016neural} needs 2000 GPU-days to obtain the final architecture by training and evaluating more than 20,000 neural architecture candidates and an EA-based method~\cite{real2019regularized} discovers the best architecture cross 3150 GPU-days. 

In this paper, we propose an \textbf{A}rchitecture-\textbf{D}riven \textbf{W}eight \textbf{P}rediction (ADWP) approach for neural architecture search (NAS), aiming to directly search for the optimal architecture rather than the best cells~\cite{liu2018darts}. As detailed in Fig~\ref{fig:space}, the search space of a cell is represented as a directed acyclic graph (DAG), which is composed of three elements: ordered nodes, colored circles and arrows.  
The DAG contains a large number of sub-graphs and each sub-graph represents a neural architecture, which forms a huge search space. Finding an optimal architecture in such a huge space is not a trivial task. To this, the search space is pruned into an intensive-space which consists of only the most likely operations.

Then, as detailed in Fig.~\ref{fig:backbone}, we leverage the intensive-space to build a HyperNetwork with a pre-defined number of cells. Moreover, each cell may contain multiple GeneratingBlocks and ConvBlocks. Each ConvBlock indicates a convolution operation in the intensive-space and the GeneratingBlock is to generate weights for the ConvBlock. To predict the weights well, the HyperNetwork is trained iteratively to drive the GeneratingBlocks converge by feeding in stochastic encoding architecture parameters. After training, target architectures with predicted weights can be evaluated efficiently without any finetuning, thus allowing us to directly search for the optimal architecture (i.e.,macro-search). Moreover, the models discovered by the proposed ADWPNAS achieve superior or comparable results.


In summary, the contribution of our paper is three-fold:

\begin{enumerate}
    \item We propose ADWPNAS, which well predicts weights for the target architectures instead of training them. This way, ADWPNAS greatly improves search efficiency, enabling us to find competitive models in few GPU-hours with macro-search.

    \item In the search procedure, we replace the original search space with an intensive-space, which is intensive in the respect of neural architectures. In this way, it significantly reduces computational cost.

    \item We obtain a series of comparable results with fewer GPU resources. On CIFAR-10, ADWPNAS is able to complete a search procedure within 4.0 GPU hours and attains comparable performance with only 1.52M parameters which is at least 40\% less than those of models discovered by the existing approaches.
\end{enumerate}

\section{Related Work}

Recently, NAS has made a significant progress. Neural architectures~\cite{liu2018progressive,liu2017hierarchical,real2017large,zoph2016neural,zoph2018learning} obtained by NAS methods have surpassed the manually designed in many fields. NAS methods can be divided into the following two categories: micro search and macro search.

\textbf{Micro search} algorithms aim to find the best neural cells, and then stack them to construct networks at different depths according to actual needs~\cite{liu2018darts,zhang2018graph,xie2018snas,dong2019searching}. Liu et al.~\cite{liu2018darts} relax the search space to continuous and then use a gradient descent method to search for the final neural cells in shallow depth. By stacking the cells into a deep network, they achieve comparable performance in the classification task. In ProxylessNAS~\cite{cai2018proxylessnas}, Cai et al. also search for the best cells and then expand on the depth of the network by stacking them together. Dong et al.~\cite{dong2019searching} search for the cells by using differentiable architecture sampler and reduce the search cost to 4 GPU-hours. One of the great advantages of micro search algorithms is that it is easy to extend in terms of depth of the network after obtaining the best cells. However, the network obtained by stacking multiple cells is probably not globally optimal.

\textbf{Macro search} algorithms, on the other hand, directly search for the optimal network instead of cells~\cite{cai2018efficient,li2019partial,veniat2018learning,chen2019progressive}. Baker et al.~\cite{baker2016designing} use reinforcement learning techniques to train the Q-Learning agent and then select the CNN layer by the agent. Based on DARTS~\cite{liu2018darts}, Chen et al.~\cite{chen2019progressive} allow the depth of the network to deepen in the search procedure, and achieve macro search to a certain extent. These traditional methods can obtain a globally optimal network in theory. However, in practice, it is inefficient and difficult to search for architectures with depths similar to ~\cite{he2016deep,huang2017densely} due to the huge search space. e.g., a depth of 12 contains $10^{29}$ possible networks\cite{pham2018efficient}. Although The proposed ADWPNAS falls into the macro search category. Different from the traditional methods, we utilize learning a HyperNetwork with reasonable weight prediction to improve search efficiency, thus enabling to search for the optimal architecture in different depths. 

\begin{figure*}[t]

\begin{center}
\includegraphics[width=1\linewidth]{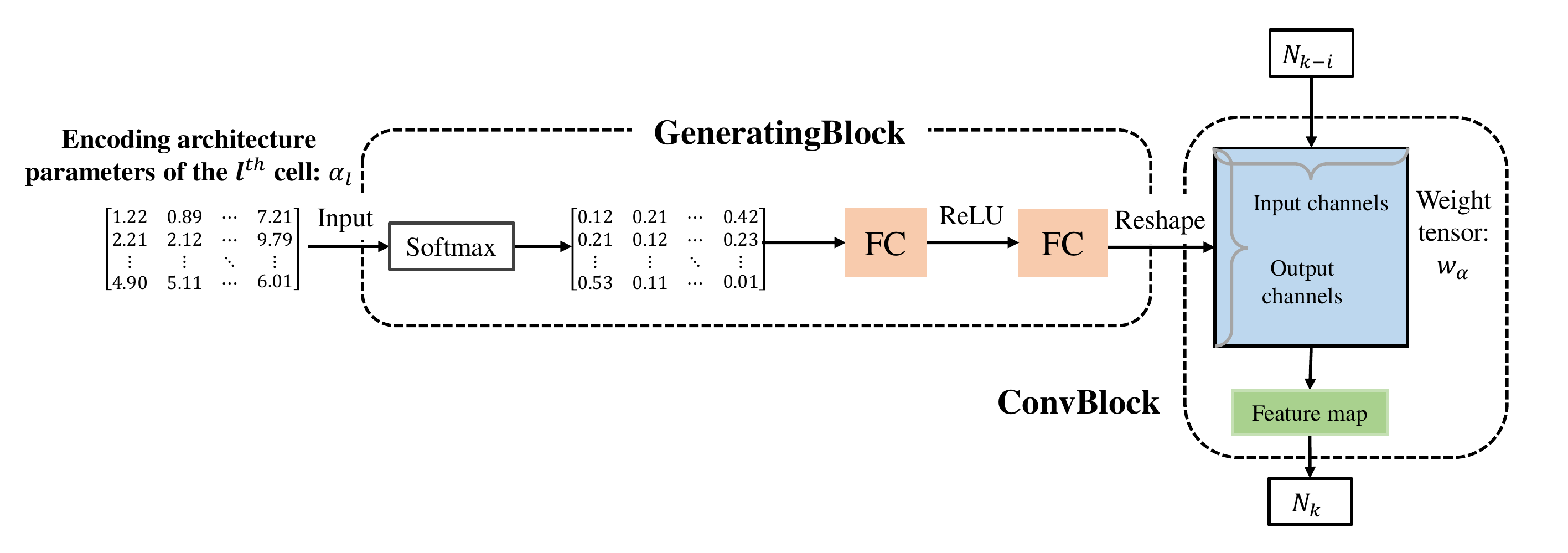}
\end{center}
    \caption{
        Network structure of GeneratingBlock connected with ConvBlock.  In the $l^{th}$ cell, ConvBlock represents a convolution operation between the nodes $N_{k-i}$ and $N_{k}$ and transform the $(k-i)^{th}$ node to an intermediate feature with the weights generated by the corresponding GeneratingBlock. Meanwhile, the GeneratingBlock takes the encoding architecture parameters $\alpha_{l}$ as input and outputs weights for the ConvBlock, as described in Sec.~\ref{subsect:training}.  
        }
\label{fig:GeneratingBlock}
\end{figure*}

\section{Methodology}
\label{sect:method}

The key challenge of NAS is how to evaluate a large number of models efficiently and accurately under resource constraints, so as to derive the optimal neural architecture. Generally, it is efficient to explore in a search space with intensive and effective architectures through a well-designed search strategy. Hence, we propose ADPWNAS to derive the optimal neural architecture in an efficient way and elaborate on two aspects: search space (Sec.~\ref{subsect:space}) and search strategy (Sec.~\ref{subsect:search_procedure}).


\subsection{Architecture Search Space}
\label{subsect:space}

In the differentiable architecture search~\cite{liu2018darts,cai2018proxylessnas,chen2019progressive}, the optimal neural architecture is usually obtained by selecting the most likely operations from distinct nodes in the search space (hereafter termed original search space). However, a sub-space, consisting of the most likely operations by selecting $K$ from the connected with all the previous nodes usually remains stable after training for a period of time (this sub-space called intensive-space below). Therefore, the architecture finally obtained is generally included in the intensive-space. This is to say, it is reasonable to derive an intensive-space from the original search space and further to obtain the optimal architecture.

\subsubsection{Original Search Space}

In this stage, we utilize the search space presented in ~\cite{zoph2016neural,zoph2018learning,liu2018darts} as our original search space to build a network of $L$ cells, but contained only normal and reduction types of cells. Each cell is represented as a DAG of $M$ nodes, $\{N_{0}, N_{1}, N_{2}, \cdots , N_{M-1}\}$, in which each node is connected with the previous by a set of operations $o(\cdot)$, $e.g.$, convolution, pooling, $zero$. Thus, all the operations in a cell constitute the original search space, denoted as $\mathcal{O}$. To make the search space continuous, a softmax function is applied to all possible operations mixed with weight $\alpha^{(i,j)}$ for each pair of node $(i,j)$:
\begin{align}
\label{equation:E1}
   & F_{i,j}(N_{i}) =
        \sum\limits_{o\in{\mathcal{O}}} 
        \frac{{\rm exp}(\alpha_{o}^{(i,j)})}
            {{\sum_{o^{'}\in{\mathcal{O}}} {\rm exp}(\alpha_{o^{'}}^{(i,j)})}}
        \ o(N_{i})                                                                 
\end{align}
where $F_{i,j}$ represents the information flow of feature maps and $\alpha^{(i,j)}$ is a vector of $|\mathcal{O}|$ dimension. Meanwhile, each node takes all the previous nodes as inputs, and outputs the sum of the features transformed from the inputs. Thus, each intermediate node can be computed as $N_{j} = \sum_{i<j}F_{i,j}(N_{i})$. Apart from this, each cell takes the outputs of the previous two cells as two input nodes ($N_{0}$ and $N_{1}$) and gets the output node $N_{M-1}$ by concatenating the intermediate nodes $\{N_{2},N_{3},\cdots,N_{M-2}\}$ in the dimension of channels. Therefore, the structure of each cell is represented as a set of variables $\{\alpha^{(i,j)}\}$. As the two types of cells share their respective mixing weights $\alpha_{normal}$ and $\alpha_{reduce}$, the architecture parameters are encoded as $\alpha=\{\alpha_{normal},\alpha_{reduce}\}$.
\subsubsection{Intensive-space Deriving}
\label{subsect:intensive-space}
As shown in Fig.~\ref{fig:space}, for each node, the most likely $K$ non-zero operations from the previous nodes are selected to form an intensive-space, denoted by $\mathcal{O}^{*}$. Assuming there are $M$ nodes (excluding two input nodes and one output node) in a cell with the original space, the cell with the intensive-space contains $M \times K$ operations. Formally, the intensive-space can be denoted by $\mathcal{O}^{*}=\{ \mathcal{O}_{i,j}^{*}|0 \leq i <j, 1 \leq j < M\}$, 
where $\mathcal{O}_{i,j}^{*}$ indicates the space consisting of a set of selected operations between the node $i$ and $j$.

On one hand, if the intensive-space is only determined based on the validation accuracy, it is likely to encounter similarities in accuracy while the corresponding architecture parameters are quite different. On the other hand, due to continuous relaxation~\cite{liu2018darts}, architecture parameters are updated continuously. Hence, the difference between the parameters $\alpha_{t}$ at the $t^{th}$ epoch and $\alpha_{t-1}$ at the last epoch are relatively small while the corresponding accuracy might be quite difference. Therefore, we define a criterion to evaluate the superiority of a search space considering both the accuracy and the stability. Specifically, the accuracy is denoted by $\epsilon(\alpha_{t})$ and the stability of an intensive-space is defined as follows:
\begin{equation}
s(\mathcal{O}_{t-i}^{*},\mathcal{O}_{t}^{*}) = 1 -
        \frac{C(\mathcal{O}_{t-i}^{*}, \mathcal{O}_{t}^{*})}
            {M \times K}
\end{equation}
where $C(\mathcal{O}^{*}_{t-i}, \mathcal{O}^{*}_{t})$ denotes the number of changed operations between the intensive-space $\mathcal{O}^{*}_{t-i}$ and $\mathcal{O}^{*}_{t}$. $M$ represents the number of nodes in an intensive-space and $K$ indicates the number of the most likely non-zero operations connected to each node.
Backtracking $n$ epochs, the superiority is expressed as follows:
\begin{equation}
\label{equation:E2}
    \mathcal{S}(\mathcal{O}_{t}^{*}) = 
        \prod\limits_{0 \leq i \leq n}
        s(\mathcal{O}_{t-i}^{*},\mathcal{O}_{t}^{*})
        \ \epsilon(\alpha_{t-i})
\end{equation}
The defined superiority has the property that it equals to $1$ when there is no change between $\mathcal{O}_{t-i}^{*}$ and $\mathcal{O}_{t}^{*}$ and the corresponding accuracy reaches $100\%$, simultaneously.

After deriving the intensive-space, it is relaxed by applying the softmax function:
\begin{align}
\label{equation:E3}
    N_{j} &= \sum \limits_{i=0}^{j-1} \sum \limits_{o \in \mathcal{O}_{i,j}^{*}}
        \frac{{\rm exp}(\alpha_{o}^{(i,j)})}
        {\sum_{i=1}^{j-1} \sum_{o^{'} \in \mathcal{O}_{i,j}^{*}} {\rm exp}(\alpha_{o^{'}}^{(i,j)})}
        o(N_{i})                                                                                \\
    s.t. & \ \ \sum \limits_{i=0}^{j-1} |\mathcal{O}_{i,j}^{*}| = K
\end{align}
where $\alpha^{(i,j)}$ is a vector of $|\mathcal{O}_{i,j}^{*}|$ dimension. Additionally, the probability of an operation is defined as $\frac{{\rm exp}(\alpha_{o}^{(i,j)})}{\sum_{i=1}^{j-1} \sum_{o^{'} \in \mathcal{O}_{i,j}^{*}} {\rm exp}(\alpha_{o^{'}}^{(i,j)})}$. 

Moreover, since there are two types of cells with the original search space, we respectively derive two types of intensive-spaces from the normal and reduction cells, which are denoted by $\mathcal{O}_{normal}^{*}$ and $\mathcal{O}_{reduce}^{*}$. Furthermore, the intensive-spaces are encoded as $\mathcal{O}^{*}=\{\mathcal{O}_{normal}^{*},\mathcal{O}_{reduce}^{*}\}$.

\begin{figure}[t]
\begin{center}
\includegraphics[width=1\linewidth]{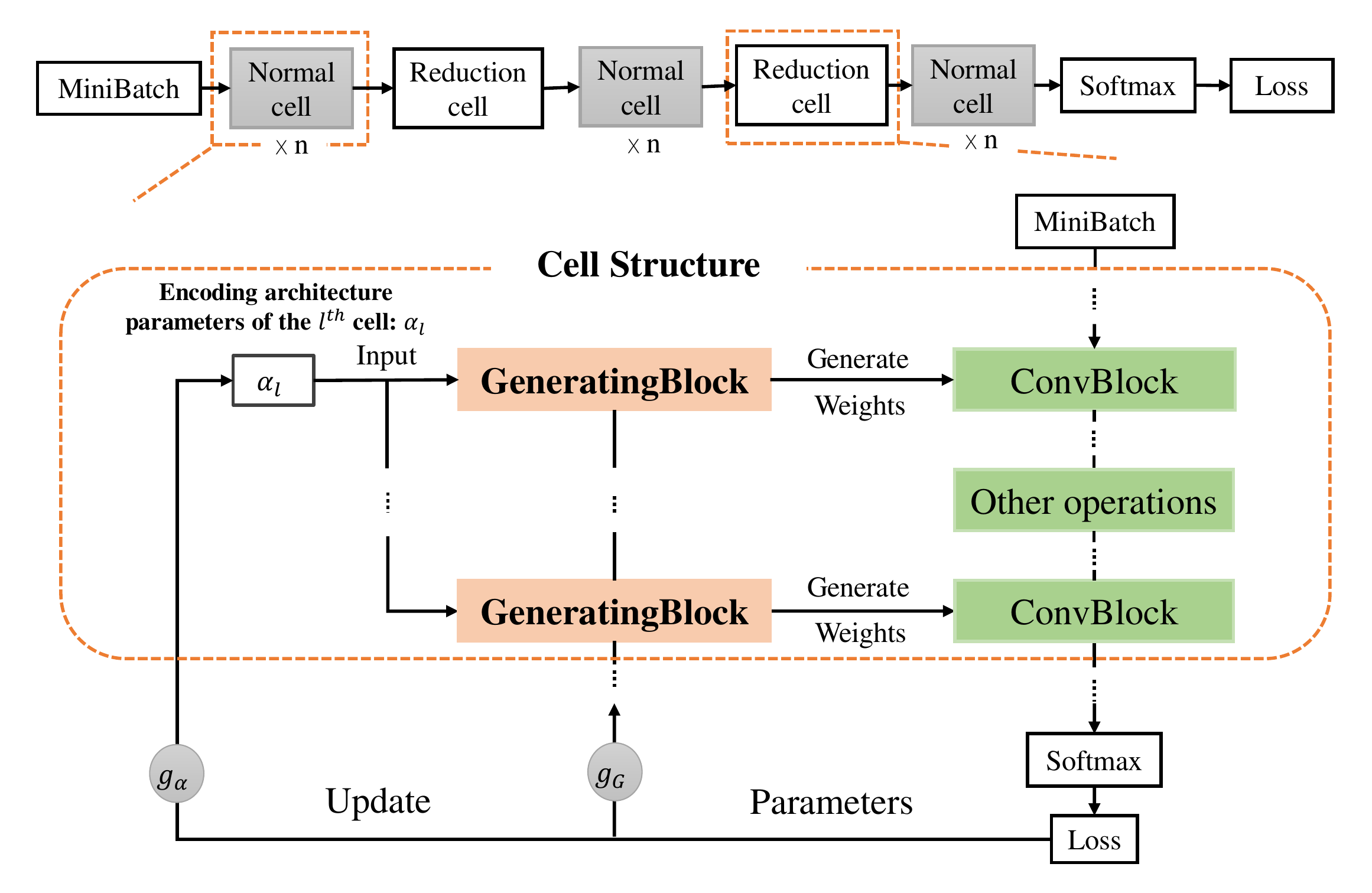}                     
\end{center}
   \caption{$Upper\ side$: The overview of HyperNetwork for ADWPNAS.
    A neural network contains a pre-defined number of cells, where reduction cells are located at 1/3 and 2/3 of the network's depth. For a minibatch of input image, $Loss$ can be calculated by the HyperNetwork with the generated weights.
    $Lower\ side$: The structure of a cell. 1) ConvBlock indicates a convolution operation and other operations consist of the operations of pooling and identity in the intensive-space.  All the cells share similar structures, in which weights of the convolution operations are generated by the GeneratingBlocks.  2) After obtaining the Loss by the HyperNetwork, we leverage gates $g_{\alpha}$ and $g_{G}$ to control whether to update the corresponding parameters or not in the training and search procedures as described in Sec.~\ref{subsect:search_procedure}.
   } 
\label{fig:backbone}
\end{figure}

\subsection{Search Procedure}
\label{subsect:search_procedure}

Let $\mathcal{L}_{train}$ and $\mathcal{L}_{val}$ represent the loss of training and validation, respectively. The optimization object of NAS is to find the optimal architecture $\alpha^{*}$ to minimize $\mathcal{L}_{val}(w^{*}, \alpha^{*})$, where $w^{*}$ indicates the weights of the network corresponding with $\alpha^{*}$. $w^{*}$ is obtained through minimizing the loss of training, expressed as $w^{*} = \argmin_{w}\ \mathcal{L}_{train}(w, \alpha^{*})$.

As pointed out in~\cite{chu2019fairnas}, since it is uncertain whether the sharing weight methods is able to reflected the real strength of architectures, we propose ADWPNAS to find the no-sharing and reasonable weights for each architecture. To be specific, the problem of NAS is divided into two consecutive sub-problems:
\begin{verse}
    1) finding the optimal weights $w^{*}_{\alpha}$ for a given architecture $\alpha$ :
    \begin{equation}
    \label{equation:E4}
        w^{*}_{\alpha} = \argmin_{w}\ \mathcal{L}_{val}(w, \alpha)
    \end{equation}
\end{verse}
\begin{verse}
    2) searching for the optimal architecture $\alpha^{*}$: 
    \begin{equation}
    \label{equation:E5}
        \alpha^{*} = \argmin_{\alpha}\mathcal{L}_{val}(w^{*}_{\alpha}, \alpha)
    \end{equation}
\end{verse}

\subsubsection{Weight Prediction and HyperNetwork Training}
\label{subsect:training}

To solve the sub-problem 1), it is straightforward to get the optimal weights by training each architecture. However, the computational cost is usually unaffordable. In this regard, we approximate the weights through prediction by building a HyperNetwork. 

The HyperNetwork is constructed in the way as shown in the upper side of Fig.~\ref{fig:backbone}, where all the normal cells share the intensive-space $\mathcal{O}_{normal}^{*}$ and the reduction ones share $\mathcal{O}_{reduce}^{*}$.
Meanwhile, architecture parameters are extended from two types of cells to the entire network. That is, each cell in the HyperNetwork has independent architecture parameters. Formally, an architecture with $L$ cells can be represented as a set of vectors $\alpha_{arch} = \{\alpha_{l}|0 \leq l < L\}$, where $\alpha_{l}$ represents a vector of parameters associated with $l^{th}$ cell. 

Furthermore, as described in Fig.~\ref{fig:GeneratingBlock}, the weights of each convolution operation in the cells are predicted by the corresponding GeneratingBlock. To be specific, a ConvBlock indicates a convolution operation between the ${k-i}^{th}$ node and the $k^{th}$ one in the $l^{th}$ cell and a GeneratingBlock is composed of a softmax function and two fully-connected layers. The GeneratingBlock takes encoding architecture parameters as input to generate the weight matrix. Then, the generated weight matrix is reshaped to match the number of input and output channel in the ConvBlock. Note that a softmax function is applied to derive the probabilities of all the operations in the $l^{th}$ cell. Moreover, there is a one-to-one correspondence between the ConvBlock and GeneratingBlock. Therefore, we formulate the process of deriving the weights $w_{\alpha}$ for the architecture with parameters $\alpha_{arch}$ in the HyperNetwork as: 
\begin{equation}
    w_{\alpha} = G(w_{G},\alpha_{arch})
\end{equation} 
where G denotes the GeneratingBlocks and $w_{G}$ represents the parameters of the GeneratingBlocks. 

For a batch of input image $x_{train}$, we can calculate the loss by the HyperNetwork with the generated weights in the ConvBlocks, detailed in the lower side of Fig.~\ref{fig:backbone}. Furthermore, we denote the process by $ Loss = ConvBlocks(x_{train})$.

With the GeneratingBlocks, finding $w^{*}$ (Equ.~\ref{equation:E4}) is transformed into solving the following $\mathcal{L}_{val}$ minimization problem:
\begin{align}
\label{equation:E6}
    G^{*} &= 
    \argmin_{G}
        \ \mathcal{L}_{val}(G(w_{G}^{*},\alpha_{arch}),\alpha_{arch})           \\
    s.t.&\ \ w^{*}_{G} =
        \argmin_{w}\mathcal{L}_{train}(G(w_{G},\alpha_{arch}),\alpha_{arch})
\end{align}


Then, the task of finding $w_{\alpha}^{*}$ for $\alpha$ reduces to optimize the GeneratingBlocks to generate weights by training the HyperNetwork.

\textbf{HyperNetwork training.} 
In the forward propagation, we change the encoding architecture parameters randomly at each iteration since the purpose is to generate weights for given architectures with different $\alpha_{arch}$.

In the backward propagation, parameters of the GeneratingBlocks rather than the ConvBlocks are updated iteratively by gradient descent. For the sake of implementation, a binary gate ($g_{G}$) is leveraged to control whether to update the parameters or not when its value equals $1$ means open and $0$ for off. 

Details are introduced in the Stage $1$ of Algorithm.~\ref{alg:alg1}.

\begin{algorithm}
\label{alg:alg1}
\caption{Search Procedure based on ADWPNAS}
\KwSty{Hyper Parameters:}
    Number of Cells: $L$,\
    Number of HyperNetwork Training Iterations: $I_{train}$,\  
    Number of Cross-search Iterations to Start: $I_{cross-search}^{start}$,\
    Number of Cross-search Iterations to End: $I_{cross-search}^{end}$,\
    Number of Total Iterations: $I_{total}$.\\
\KwIn{intensive-space:\ $\mathcal{O}^{*}$, training dataset: $x_{train}$, validation dataset: $x_{val}$.}
\KwOut{The\ Discrete\ Architecture\ Obtained.}\ \\
\KwSty{Stage\ 1.\ HyperNetwork\ training:}\\
    \For{i=0\ :\ $I_{train}$}
    {
        Generate architecture parameters $\alpha_{arch}$ randomly;\\
        Obtain $w_{\alpha}$ by $GeneratingBlocks(\alpha_{arch})$;\\
        Obtain $Loss$ by $ConvBlocks(x_{train})$;\\
        Update the parameters of GeneratingBlocks $w_{G}$;\\
    }\ 
    Return the trained GeneratingBlocks $G^{*}$.\\ \ \\

\KwSty{Stage\ 2.\ Architecture\ search:}\\
    Initialize architecture parameters $\alpha_{arch}$ randomly;\\
    \For{i=$I_{train}$\ :\ $I_{total}$ }
    {
        Obtain $w_{\alpha}$ by $GeneratingBlocks(\alpha_{arch})$;\\
        Obtain $Loss$ by $ConvBlocks(x_{train})$;\\
        Update architecture parameters $\alpha_{arch}$ by gradient descent;\\
        \If{$I_{cross-search}^{start} \leq  i \textless I_{cross-search}^{end}$ }{
            Update the parameters of GeneratingBlocks $w_{G}$;\\
        }
    }\
    Obtain the best architecture parameters $\alpha^{*}_{arch}$ and return the discrete architecture.

\end{algorithm}

\begin{table*}[t]
\begin{center}

\begin{tabular}{|l|c|c|c|c|c|}
\hline
\multirow{2}{*}{\textbf{Method}} &
     \textbf{Test Error}    & \textbf{Params}   & \textbf{Search Cost}  & \textbf{Search }  \\
    & \textbf{(\%)}         & \textbf{(M)}      & \textbf{(GPU-days)}   & \textbf{Method}   \\
\hline\hline
DenseNet-BC~\cite{huang2017densely} & 3.46 & 1.7 & -- & Manual \\
\hline
NASNet-A~\cite{zoph2018learning} + cutout       & 2.65  & 3.3   & 1800  & RL \\
AmoebaNet-A + cutout~\cite{real2019regularized} & 3.34  & 3.2   & 3150  & Evolution\\
AmoebaNet-B + cutout~\cite{real2019regularized} & 2.55  & 2.8   & 3150  & Evolution \\
PNAS~\cite{liu2018progressive}                  & 3.41  & 3.2   & 225   & SMBO \\
ENAS + cutout~\cite{pham2018efficient}          & 2.89  & 4.6   & 0.5   & RL \\
DARTS ((first order) + cutout~\cite{liu2018darts}        & 3.00  & 3.3   & 1.5   & Gradient \\
DARTS (second order) + cutout~\cite{liu2018darts}        & 2.76  & 3.3   & 4.0   & Gradient \\
SNAS + moderate constraint + cutout~\cite{xie2018snas}      & 2.85  & 2.80  & 1.5   & Gradient \\
GDAS + cutout~\cite{dong2019searching}          & 2.93  & 3.40  & 0.21  & Gradient \\
GDAS(FRC) + cutout~\cite{dong2019searching}     & 2.82  & 2.50  & 0.17  & Gradient \\
\hline
ADWPNAS(8-layers) + cutout                         & 2.77  & 1.52  & 0.2     & Gradient \\    
ADWPNAS(8-layers) + cutout + AutoAugment           & \textbf{2.41}  & \textbf{1.52}  & 0.2     & Gradient \\    
ADWPNAS(14-layers) + cutout                        & 2.55  & 2.62  & 0.4     & Gradient \\      
ADWPNAS(14-layers) + cutout + AutoAugment          & \textbf{2.08}  & \textbf{2.62}  & 0.4     & Gradient \\
\hline
\end{tabular}
\end{center}
\caption{Comparison with state-of-the-art architectures on CIFAR-10. The search costs are derived from the original papers. Note that the search cost for ADWPNAS include the HyperNetwork training cost and the architecture search cost, but exclude the intensive-space deriving cost(0.34 GPU days). Our experiments are based on the TiTAN RTX GPU.}
\label{table:exp-cifar}
\end{table*}

\subsubsection{Architecture Search}
\label{subsect:search}

After training the HyperNetwork, the optimization object (Equ.~\ref{equation:E5}) can be transformed into the following by $G^{*}(w_{G}^{*},\alpha_{arch})$:
\begin{align}
\label{equation:E8}
  & \alpha_{arch}^{*} = 
    \argmin_{\alpha_{arch}^{'}}
        \ \mathcal{L}_{val}(G^{*}(w_{G}^{*}, \alpha_{arch}^{'}),\alpha_{arch}^{'})           \\
  & s.t.\ \ \alpha_{arch}^{'} =
        \argmin_{\alpha_{arch}}\mathcal{L}_{train}(G^{*}(w_{G}^{*}, \alpha_{arch}),\alpha_{arch})
\end{align}
To obtain a discrete architecture, we retain the top-$T$ most likely operations (connected to each node from all the previous) according to $\alpha_{arch}^{*}$ and set $T=2$ following the existing works~\cite{liu2018darts}. 

Since the weights of convolution kernels can be derived through prediction instead of training, target architectures can be evaluated efficiently without any finetuning, which enables us to directly search for the optimal architecture (i.e., macro-search). 
As shown in Algorithm.~\ref{alg:alg1}, different from the stage of HyperNetwork Training, encoding architecture parameters are updated by gradient descent rather than being randomly generated at each iteration. Meanwhile, the binary gate $g_{\alpha}$ is on while $g_{G}$ is off. We refer to the process as the basic-search.

The basic-search is able to guide the architecture parameters $\alpha_{arch}$ to converge, though. For the converged region of $\alpha_{arch}$, there may exist some noise in the process of predicting weights owing to training the HyperNetwork randomly. To deal with the problem, cross-search is adopted when $\alpha_{arch}$ is converging in the basic-search. In the cross-search, $g_{G}$ is open as well as $g_{\alpha}$, allowing to update the weights of all the GeneratingBlocks and the architecture parameters simultaneously. After the cross-search, close $g_{G}$ and perform basic-search again to obtain the architecture parameters $\alpha_{arch}^{*}$. Therefore, the optimal architecture can be derived from $\alpha_{arch}^{*}$.


\section{Experiments and Results}

Our experiments on CIFAR-10~\cite{krizhevsky2009learning} consist of three parts, intensive-space deriving(Sect.~\ref{subsect:exp-intensive-space}), search procedure(Sec.~\ref{subsect:exp-search}) and architecture evaluation(Sec.~\ref{subsect:exp-evaluation}). 
Additionally, the transferability of the architectures learned on CIFAR-10 is investigated by evaluating them on ImageNet~\cite{russakovsky2015imagenet}.

\begin{table*}[t]
\begin{center}
\begin{tabular}{|l|c|c|c|c|c|c|}
\hline
\multirow{2}{*}{\textbf{Method}}
    & \multicolumn{2}{c|}{\textbf{Test Error(\%)}} & \textbf{Params} & $+$$\times$ & \textbf{Search Cost}  & \textbf{Search }  \\
      \cline{2-3}
    & \textbf{Top-1} & \textbf{Top-5}              & \textbf{(M)}    & \textbf{(M)}  & \textbf{(GPU-days)}   & \textbf{Method}   \\
\hline\hline
Inception-v1~\cite{szegedy2015going} & 30.2 & 10.1 & 6.6 & 1448 & -- & Manual \\
MobileNet~\cite{howard2017mobilenets} & 29.4 & 10.5 & 4.2 & 569 & -- & Manual \\
\hline
NASNet-A~\cite{zoph2018learning}       & 26.0  & 8.4 & 5.3 & 564 & 1800  & RL \\
AmoebaNet-A~\cite{real2019regularized} & 25.5  & 8.0 & 5.1 & 555 & 3150  & Evolution \\
AmoebaNet-B~\cite{real2019regularized} & 26.0  & 8.5 & 5.3 & 555 & 3150  & Evolution \\
AmoebaNet-C~\cite{real2019regularized} & 24.3  & 7.6 & 6.4 & 570 & 3150  & Evolution \\
PNAS~\cite{liu2018progressive}         & 25.8  & 8.1 & 5.1 & 588 & 225   & SMBO \\
DARTS (second order)~\cite{liu2018darts}        & 26.7  & 8.7 & 4.7 & 574 & 4.0   & Gradient \\
SNAS + moderate constraint~\cite{xie2018snas}      & 27.3  & 9.2 & 4.3 & 522 & 1.5   & Gradient \\
GDAS~\cite{dong2019searching}          & 26.0  & 8.5 & 5.3 & 581 & 0.21  & Gradient \\
GDAS(FRC)~\cite{dong2019searching}     & 27.5  & 9.1 & 4.4 & 497 & 0.17  & Gradient \\
\hline
ADWPNAS(8-layers)                         & \textbf{27.6}  & 9.4 & \textbf{3.7} & \textbf{389} & 0.2     & Gradient \\   
ADWPNAS(14-layers)                        & 26.4  & 8.6 & 5.3 & 565 & 0.4     & Gradient \\
\hline
\end{tabular}
\end{center}
\caption{Comparison with state-of-the-art architectures on ImageNet (mobile setting). $+$$\times$ indicates the number of multiply-add operations.}
\label{table:exp-imagenet}
\end{table*}

\subsection{Intensive-space Deriving}
\label{subsect:exp-intensive-space}

In the proposed ADWPNAS, our original search space consists of the following 8 operations: $3\times3$ depthwise-separable conv, $5\times5$ depthwise-separable conv, $3\times3$ dilated-separable conv, $5\times5$ dilated-separable conv, $3\times3$ average pooling, $3\times3$ max pooling, identity and $zero$. 

The backbone network is constructed by stacking $L$ cells and each cell contains $M=7$ nodes. The inputs are the first and second nodes of the $l^{th}$ cell, which equal to the outputs of the ${(l-2)}^{th}$ and ${(l-1)}^{th}$ cells respectively, and the output is set to be the ${6}^{th}$ node. Reduction cells, detailed in the upper side of Fig.~\ref{fig:backbone}, are located at $1/3$ and $2/3$ of the network's depth, which are connected to normal cells by operations with stride of two. 

We search for two intensive-spaces\footnote{The intensive-spaces are provided in the supplementary material.}, denoted by $\mathcal{O}_{8}^{*}$ and $\mathcal{O}_{14}^{*}$, which are obtained under the configuration of 8 cells and 14 cells with $K=6$ and $n=2$, respectively. The other experiment settings are same as DARTS, except that batch size is set to be 200 and the number of epochs for training is 30. 

\begin{figure}[t]
\begin{center}
   \includegraphics[width=0.8\linewidth]{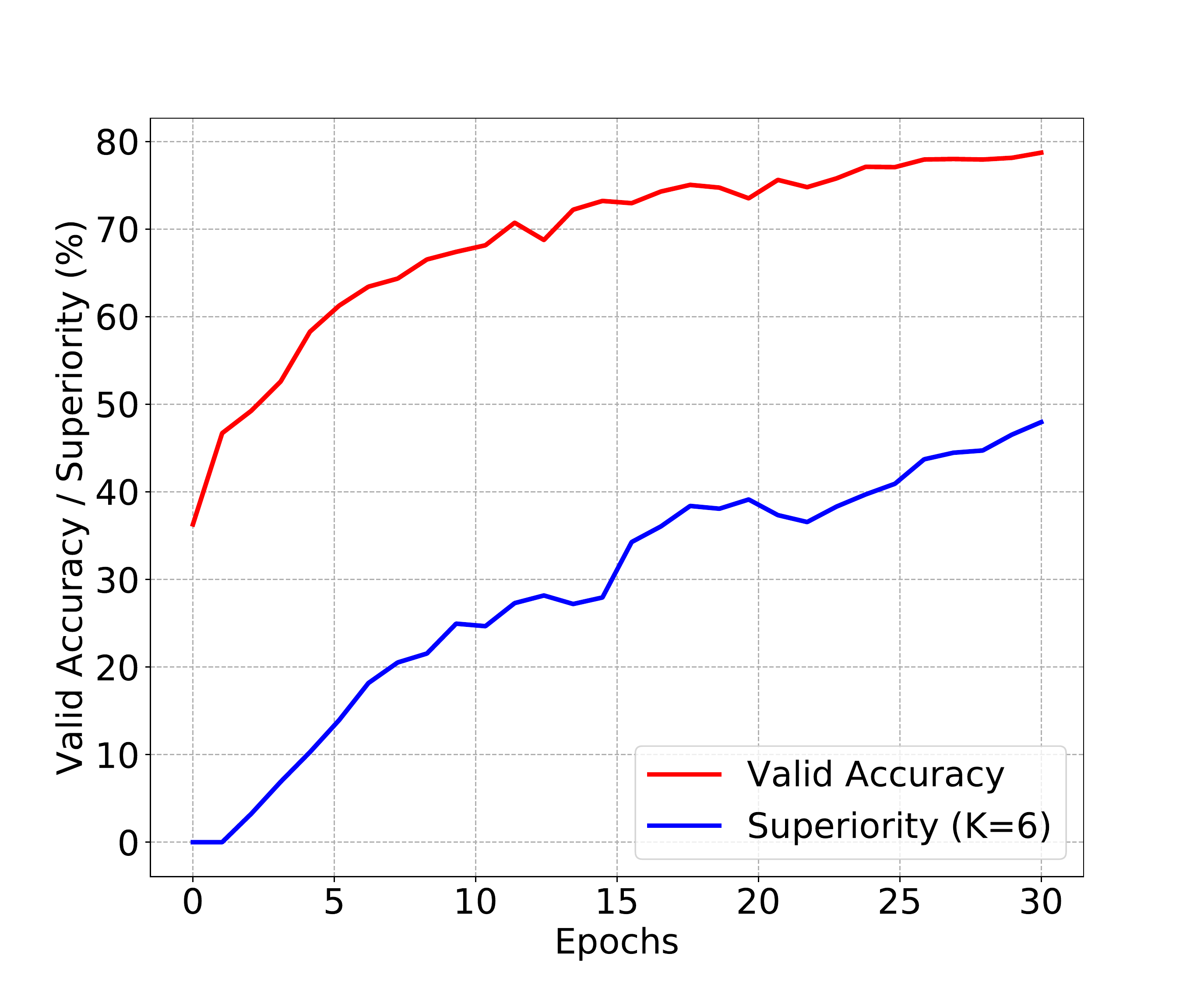}
\end{center}
   \caption{This figure presents the validation accuracy and superiority of an intensive-space with $K=6$.}
\label{fig:exp-S}
\end{figure}

\textbf{Discussion about deriving the intensive-space.} Fig.~\ref{fig:exp-S} illustrates the validation accuracy and the superiority of an intensive-space ($K=6$). Although the two have similar trends, the superiority shows significant differences at different times (epochs) while the validation accuracy gradually converges. This result indicates that the superiority is helpful to identify the performance of intensive-spaces.


\subsection{Search Procedure}
\label{subsect:exp-search}

Our search procedure is composed of two stages. In the first stage, the HyperNetwork is randomly trained from scratch. In the second stage, we search the best architecture based on the HyperNetwork by gradient descent. However, both the two stages share the same backbone network, detailed in Fig.~\ref{fig:backbone}. 
In a GeneratingBlock of the $l^{th}$ cell, we use the encoding architecture parameters $\alpha_{l}$ as input and apply a softmax function to calculate the probabilities of all the operations in the cell. Then, the probability (associated with the operation) is used as input of the first full-connected layer, which outputs a vector with the size of 64. The second full-connected layer utilizes the 64-size vector as input to generate a matrix in shape of $(1,c_{l}^{out} \times c_{l}^{in} \times w_{l} \times h_{l})$. Immediately, the output matrix is reshaped to $(c_{l}^{out} \times c_{l}^{in} \times w_{l} \times h_{l})$ as the weights, where $c_{l}^{in}$ and $c_{l}^{out}$ respectively indicate the number of input and output channel.



In the procedure, we search for architectures with 98 batch size under the configuration of $I_{train}=60$, $I_{cross-search}^{start}=90$, $I_{cross-search}^{end}=100$ and $I_{total}=120$. For $w_{G}$, we leverage the SGD~\cite{polyak1992acceleration} optimizer with momentum $\beta = 0.9$ and weight decay 3e-4. The learning rate is initialized to 0.025 and annealed down to 0 following a cosine schedule. For $\alpha_{arch}$, they are generated randomly at each iteration in the HyperNetwork training stage. In the search stage, however, $\alpha_{arch}$ are optimized by the Adam~\cite{kingma2014adam} optimizer with momentum $\beta =(0.5, 0.999)$, learning rate 3e-4 and weight decay 1e-3. Additionally, other experimental settings are the same as ~\cite{liu2018darts}. 

Note that $\mathcal{O}_{8}^{*}$ and $\mathcal{O}_{14}^{*}$ are leveraged to search for architectures\footnote{The architectures are provided in the supplementary material.} with 8 layers and 14 layers, respectively. 

\begin{figure}[t]
\begin{center}
   \includegraphics[width=0.8\linewidth]{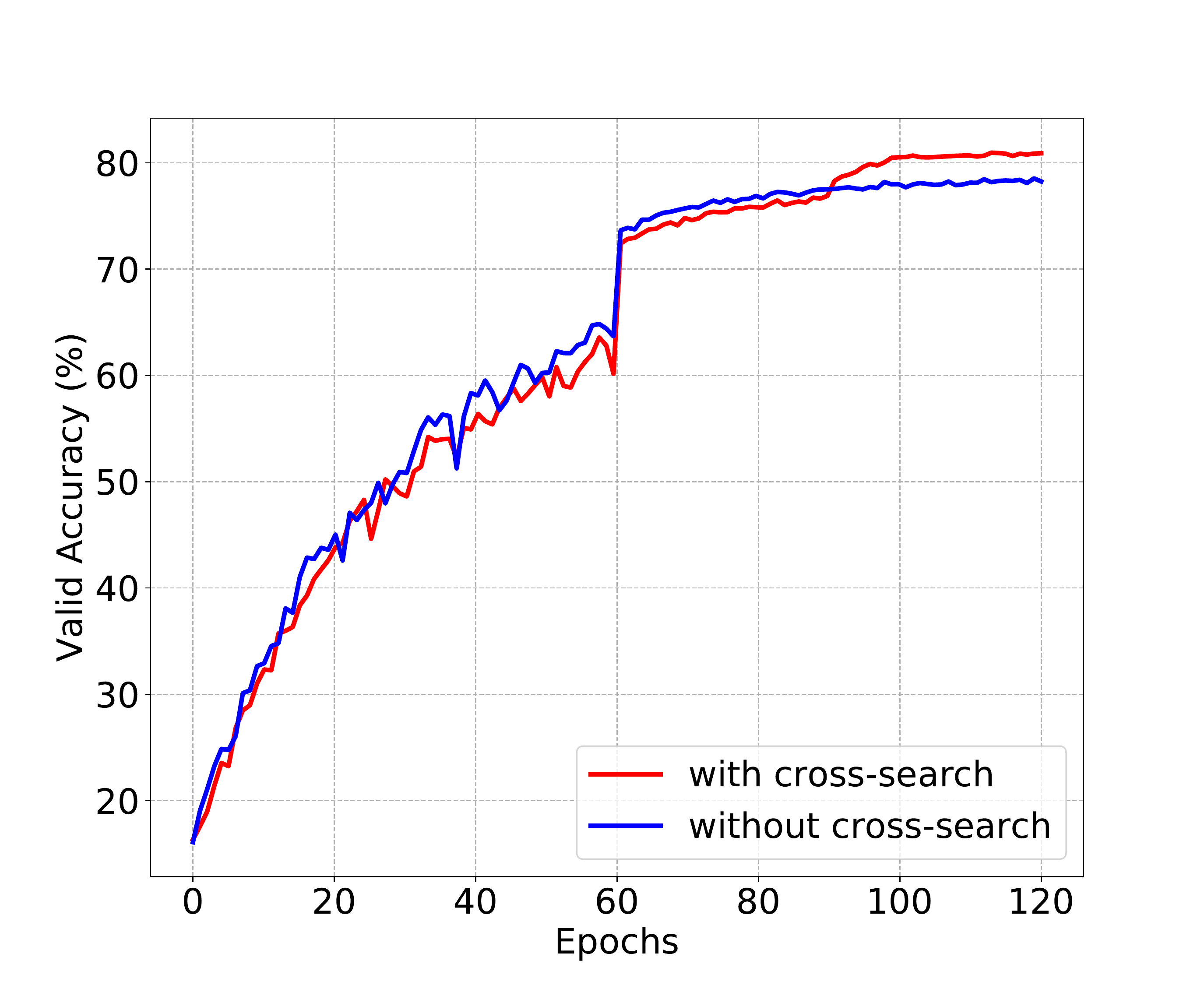}
\end{center}
   \caption{The comparison chart about validation accuracy whether or not the cross-search is included in the search process.}
\label{fig:exp-search}
\end{figure}

\textbf{Discussion about the cross-search.} 
As illustrated in Fig.~\ref{fig:exp-search}, the red line indicates the validation accuracy in the search procedure with cross-search while the blue one indicates the accuracy without cross-search. Note that the red line describes the acquistion process for the 14-layer model, and the experimental configuration of the bule line is the same with the red one expect $I_{cross-search}^{start}=I_{cross-search}^{end}=120$. 

Obviously, the red line has higher accuracy than blue after cross-search. Forasmuch, it is critical for cross-search to improve the accuracy of the search procedure and eliminate the noise of training HyperNetwork randomly.

\textbf{Complexity analysis.} We analyze the complexity of the intensive-space for neural architectures in the search procedure. Without considering graph isomorphism, there are $(\frac{K!}{T! (K-T)!})^{M-3}$ possible sub-graphs contained in each of our discretized cell (recall that there are two input and one output nodes). Since we learn each cell to derive the final architecture, the total number of architecture is up to $(\frac{K!}{T! (K-T)!})^{(M-3) \times L}$ when there are $L$ cells in the HyperNetwork. Therefore, before discretization, the continuous spaces of the HyperNetworks with 8 and 14 cells cover $(\frac{6 \times 5}{2})^{4 \times 8} \approx 10^{37}$ and $(\frac{6 \times 5}{2})^{4 \times 14} \approx 10^{65}$ architectures, respectively. Those two are far greater than $10^{25}$ of DARTS.

\subsection{Architecture Evaluation}
\label{subsect:exp-evaluation}


\textbf{Evaluation on CIFAR-10.}
After the search procedure, we evaluate the final architectures by training from scratch on CIFAR-10 and report its accuracy on the test set. For training only with the standard cutout~\cite{devries2017improved} trick, we follow the same settings as ~\cite{liu2018darts}, but for 1000 epochs with batch size 98 to converge. When adding the AutoAugment~\cite{cubuk2018autoaugment} technology, the number of epochs increases to 1500 and the other settings remain the same.

The comparison between the models discovered by ADWPNAS and other state-of-the-art models are summarized in Tab. ~\ref{table:exp-cifar}. The 8-layer model discovered by our approach not only achieves the test error rate of 2.77\% on CIFAR-10, but also reduces the parameters to 1.52M, which is 40\% less than those of other models with the same level of accuracy. When adding AutoAugment trick, it achieves a lower error rate of 2.41\%. Furthermore, the 14-layer model  with more parameters can achieve better results, which are test error rates of 2.55\% and 2.08\% (with AutoAugment trick). However, the parameters contained in the model are also less than those of other state-of-art models.

\textbf{Evaluation on ImageNet.}
The experimental setup on ImageNet is exactly the same as DARTS but with batch size of 256 and auxiliary weight of 0.7, and the experimental results are shown in Tab.~\ref{table:exp-imagenet}. Meanwhile, ADWPNAS(8-layers) trained with initial channel size 50 saves 16\% less model parameters aside with 20\% less multiply-add operations than GDAS(FRC)~\cite{dong2019searching}, and achieves a similar top-1 error rate of 27.6\%. Furthermore, we successfully transfer the 14-layer model to ImageNet with competitive performance.


\section{Conclusion}
In this paper, we propose an architecture-driven weight prediction approach for neural architecture search, which is efficient and reduces the search cost by about $10^{4}$ times compared to the NAS approach~\cite{zoph2018learning}. Moreover, the model discovered by our ADWPNAS can achieve comparable results with less parameters, especially on the dataset of CIFAR-10.

{\small
\bibliographystyle{ieee_fullname}
\bibliography{egbib}
}

\end{document}